\documentclass[sigconf, nonacm]{acmart}

\setcopyright{none}
\makeatletter
\renewcommand\@acmDOI{}
\renewcommand\@acmISBN{}
\makeatother
\usepackage{placeins}
\usepackage{microtype,graphicx,subfigure}
\usepackage{booktabs,multirow,hyperref}
\usepackage{algorithm,algorithmic}

\usepackage{amsmath,amssymb,amsthm,mathtools}

\usepackage{tikz,pgfplots,xcolor}
\usepackage{makecell}
\usepackage[capitalize,noabbrev]{cleveref}
\usepackage[textsize=tiny]{todonotes}
\usepackage[most,skins,theorems]{tcolorbox}

\pgfplotsset{compat=1.18}
\usepgfplotslibrary{fillbetween}
\usepgfplotslibrary{groupplots}
\definecolor{darkgreen}{RGB}{0,100,0}
\definecolor{darkblue}{RGB}{0,0,139}

\theoremstyle{plain}

\theoremstyle{definition}

\theoremstyle{remark}

\tcbset{
  aibox/.style={
    width=\linewidth,
    top=7pt,
    bottom=2pt,
    colback=blue!6!white,
    colframe=black,
    colbacktitle=black,
    enhanced,
    center,
    before skip=40pt,
    after skip=40pt,
    attach boxed title to top left={yshift=-0.1in,xshift=0.15in},
    boxed title style={boxrule=0pt,colframe=white,},
  }
}

\title{Optimizing Donor Outreach for Blood Collection Sessions: A Scalable Decision Support Framework}

\author{André Carneiro}
\authornote{Corresponding author: andrejoaocarneiro@tecnico.ulisboa.pt}
\affiliation{%
  \institution{INESC-ID and Técnico\\ Universidade de Lisboa}
  \country{Portugal}}

\author{Pedro T. Monteiro}
\affiliation{%
  \institution{INESC-ID and Técnico\\ Universidade de Lisboa}
  \country{Portugal}}

\author{Rui Henriques}
\affiliation{%
  \institution{INESC-ID and Técnico\\ Universidade de Lisboa}
  \country{Portugal}}
  
\begin{document}
\sloppy

\begin{abstract}
Blood donation centers face challenges in matching supply with demand while managing donor availability. Although targeted outreach is important, it can cause donor fatigue via over-solicitation. Effective recruitment requires targeting the right donors at the right time, balancing constraints with donor convenience and eligibility. Despite extensive work on blood supply chain optimization and growing interest in algorithmic donor recruitment, the operational problem of assigning donors to sessions across a multi-site network, taking into account eligibility, capacity, blood-type demand targets, geographic convenience, and donor safety, remains unaddressed.

We address this gap with an optimization framework for donor invitation scheduling incorporating donor eligibility, travel convenience, blood-type demand targets, and penalties. We evaluate two strategies: (i) a binary integer linear programming (BILP) formulation and (ii) an efficient greedy heuristic. Evaluation uses the registry from Instituto Português do Sangue e da Transplantação (IPST) for invite planning in the Lisbon operational region using 4-month windows. A prospective pipeline integrates organic attendance forecasting, quantile-based demand targets, and residual capacity estimation for forward-looking invitation plans. Results reveal its key role in closing the supply–demand gap in the Lisbon operational region. A controlled comparison shows that the greedy heuristic achieves results comparable to the BILP, with 188$\times$ less peak memory and 115$\times$ faster runtime; trade-offs include 3.9\,pp lower demand fulfillment (86.1\% vs.\ 90.0\%), larger donor--session distance, higher adverse-reaction donor exposure, and greater invitation burden per non-high-frequency donor, reflecting local versus global optimization. Experiments assess how constraint-aware scheduling can close gaps by mobilizing eligible inactive/lapsing donors.
\end{abstract}

\maketitle

\section{Introduction}
Blood donation centers worldwide face a fundamental operational challenge: maintaining adequate blood supplies to meet clinical demand while managing a voluntary donor base with limited availability and susceptibility to recruitment fatigue. In Portugal, the Instituto Português do Sangue e da Transplantação (IPST) has issued repeated public appeals as blood reserves frequently fall below safe operational thresholds \cite{theportugalnewsBloodDonationsNeeded2025, institutoportuguesdosangueedatransplantacaoipstReservasSangue, euronewsMajorHospitalPortugal09:55:41+02:00}. These shortages pose direct risks to patient care. Blood components have critically short shelf lives, with platelets expiring within five to seven days and red blood cells within thirty-five to forty-two days, making rapid, responsive supply management essential.

Traditional donor mobilization strategies rely heavily on broad, untargeted outreach campaigns: mass SMS messages, general telephone appeals, or public media announcements that contact large donor populations indiscriminately. While personal outreach, particularly phone calls, consistently produces strong attendance effects for specific sessions and for reactivating inactive donors \cite{ou-yangEffectiveMethodsReactivating2020, porto-ferreiraRandomizedTrialEvaluate2017, bruhinCallDutyEffects2015}, such approaches can become inefficient and counterproductive at scale. Over-solicitation leads to donor fatigue, with repeated contacts desensitizing recipients and increasing opt-out rates \cite{bruhinCallDutyEffects2015, domaine2010_donor_manual}. This creates a fundamental paradox: blood banks must increase outreach to address shortages, yet risk alienating their donor base through excessive solicitation.

European best-practice guidance recommends targeted, segment-specific communications with appropriate ``silence periods'' between invitation attempts \cite{ou-yangEffectiveMethodsReactivating2020, europeancommission.jointresearchcentre.BloodDonationEU2024, mohammedStratificationBloodDonors2025}. High-performing blood centers emphasize precision over volume, focusing on recruiting ``the right donors at the right time for the right collection locations'' rather than maximizing sheer contact counts \cite{5BestPractices}. Recent empirical evidence supports this approach: machine-learning-based selection of highly motivated donors achieved approximately 1.8 times higher turnout compared to conventional random or sequential invitation methods \cite{wuPredictingWillingnessDonate2022}. However, translating these insights into operational practice requires solving complex combinatorial assignment problems that simultaneously satisfy medical eligibility rules, geographic convenience constraints, and blood type demand requirements.

\begin{figure*}[!t]
    \centering
    \includegraphics[width=0.8\linewidth]{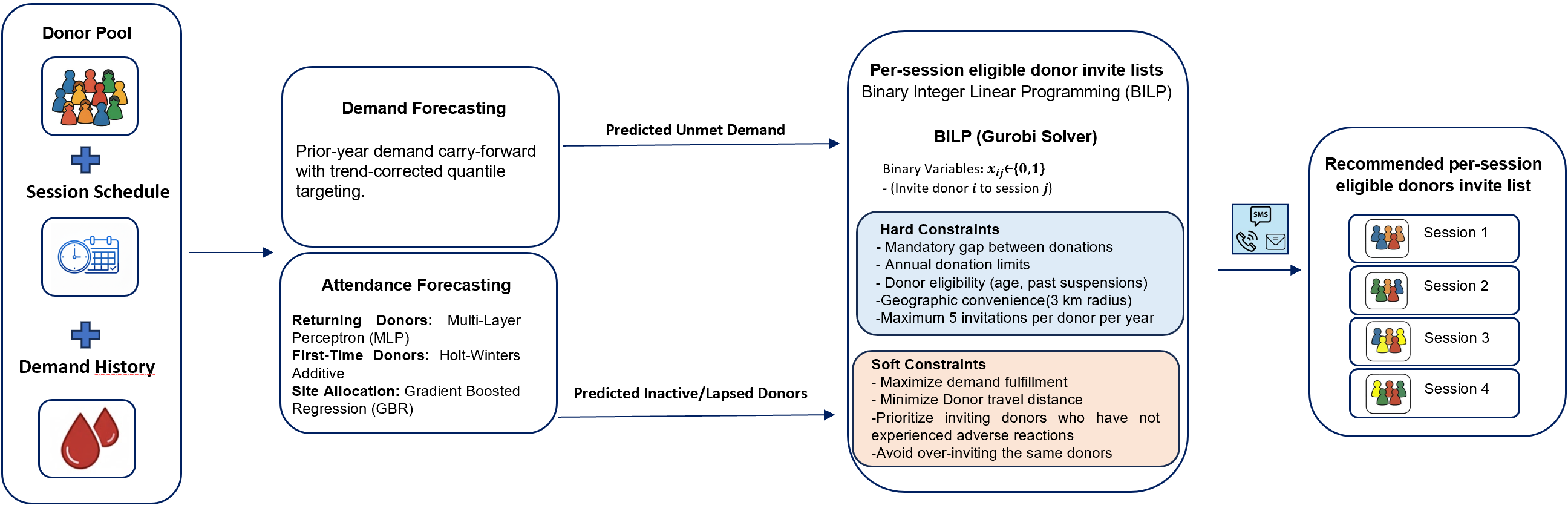}
    \caption{Strategic donor outreach framework enabled by binary integer linear programming.}
    \label{fig:example}
\end{figure*}

Current approaches to donor recruitment remain limited \cite{osorioStructuredReviewQuantitative2015, basgureUnaddressedProblemsResearch2018, mcelfreshMatchingAlgorithmsBlood2023}. 
Existing literature focuses primarily on donor motivation and behavioral interventions \cite{domaine2010_donor_manual, europeancommission.jointresearchcentre.BloodDonationEU2024}, donor classification and segmentation strategies \cite{mohammedStratificationBloodDonors2025}, prediction models for donor return probability \cite{wuPredictingWillingnessDonate2022}, or online platform-based notification matching, which frames donor recruitment as an online bipartite matching problem to optimize which donation opportunity to surface to registered users \cite{mcelfreshMatchingAlgorithmsBlood2023}. Although these approaches address important aspects of donor recruitment, they provide limited guidance on the operational problem of assigning specific donors to specific collection sessions at scale, jointly enforcing medical eligibility, session capacity, blood-type demand, and geographic constraints. The computational challenges of this problem involve hundreds of thousands of donors, hundreds of sessions, and millions of potential assignments under complex temporal constraints have not been adequately addressed in previous work.

From a computational perspective, targeted invitation scheduling is an exact combinatorial optimization problem \cite{conforti2014integer, nemhauser1988integer}. A natural modeling choice is a \emph{binary integer linear program}, where binary decision variables encode donor--session invitations and linear constraints capture capacity, demand, and feasibility rules. Modern ILP solvers typically rely on \emph{linear programming} (LP) relaxations combined with \emph{branch-and-bound} and \emph{cutting-plane} separation (often referred to as branch-and-cut), and they allow callback-based enforcement where certain constraints are generated only when violated (\emph{lazy constraints}) \cite{ParameterReferenceGurobi,pearceGeneralFormulationLazy2019}. Complementary exact paradigms such as \emph{SAT/MaxSAT} encode feasibility and preferences as Boolean formulas and rely on conflict-driven clause learning, which can be particularly effective when constraints are predominantly logical. In this work, We adopt a binary integer linear programming (BILP) formulation because it directly supports linear demand-cost trade-offs and readily accommodates the complex temporal eligibility rules governing donor availability \cite{conforti2014integer, padberg1991branchcut, biere2021handbook}.

To address these gaps, we focus on the following central research questions:
\begin{itemize}
\item \textit{Can a BILP formulation provide a useful exact framework for improving donor--session assignments in real-world blood collection systems?}
\item \textit{How does an efficient greedy heuristic compare to the exact BILP benchmark in terms of solution quality, runtime, and peak memory on realistic instances?}
\item \textit{What operational insights emerge when combining organic attendance forecasting with a constraint-aware invitation pipeline in a retro planning setting?}
\end{itemize}

In this work, we present a practical decision support framework for strategic donor invitation scheduling (Figure~\ref{fig:example}). We formulate the donor--session assignment problem as a BILP with hard constraints on medical eligibility, geographic proximity, collection site capacity, and blood-type demand, and augment it with a penalty for inviting donors with prior adverse reactions. To support operational deployment on large-scale instances and compute constrained scenarios, we also introduce a greedy heuristic that provides a fast, high-quality alternative, benchmarked against the exact BILP formulation.

We apply this framework to historical data from IPST's Lisbon zone operations, demonstrating its feasibility on real-world data. The selected study case is characterized by significant operational challenges, including persistent supply-demand imbalances where monthly blood consumption exceeds collection, while showing that constraint-aware outreach has the potential to substantially improve recruitment efficiency by ensuring targeted, convenient invitations.

The primary contributions of this work are:
\begin{enumerate}
  \item A \emph{BILP} formulation of the donor--session assignment problem incorporating medical, geographic, demand, and adverse-reaction safety constraints;
  \item A greedy heuristic (\emph{Greedy}) that prioritizes non-adverse-reaction donors and is empirically validated against the exact BILP benchmark at a fraction of the computational cost;
  \item A fully prospective invitation pipeline combining organic attendance forecasting and quantile-based demand targets;
  \item Application to real-world data from Portuguese blood services and empirical evidence of chronic supply--demand imbalances in metropolitan blood collection.
\end{enumerate}

\section{Decision Support Framework}
\label{sec:opt_framework}

This section introduces the proposed BILP formulation for the donor-session invitation assignment problem and describes the computational techniques employed to solve large-scale instances.

\subsection{Problem Formulation}
\label{sec:opt_formulation}

We model invitation planning over a horizon containing multiple collection sessions. Let $\mathcal{D}$ denote the set of candidate donors and $\mathcal{S}$ the set of scheduled sessions. For each feasible donor--session pair $(i,j)$, we define a binary decision variable
\begin{equation}
x_{ij} \in \{0,1\},
\end{equation}
where $x_{ij}=1$ indicates that donor $i$ is invited to session $j$.

\paragraph{Feasible assignment set.}
Variables are created only for donor--session pairs that pass a static pre-filtering stage. Let
\[
\mathcal{A} \subseteq \mathcal{D}\times\mathcal{S}
\]
denote the feasible pair set after checking: (i) age eligibility at the session start date, (ii) absence of an active suspension at the session start date, (iii) a minimum gap from the donor's most recent historical donation, and (iv) geographic feasibility under a maximum travel radius.

\paragraph{Expected-attendance weighting.}
Each donor $i$ is assigned an expected attendance probability $p_i \in (0,1]$, derived from the donor-type category used in the operational pipeline. Thus, inviting donor $i$ contributes $p_i$ expected donations toward monthly demand fulfillment and consumes $p_i$ units of expected session capacity.

\paragraph{Distance cost.}
Because explicit donor location preferences are not available in the operational registry, we approximate travel convenience using up to two geographic anchor points for each donor: (1) the approximate location of their most recently attended collection brigade, and (2) the approximate coordinates of their home postal code. For a feasible pair $(i,j)$, let
\[
\delta_{ij} = \min\{\mathrm{dist}^{(1)}_{ij},\mathrm{dist}^{(2)}_{ij}\}
\]
be the minimum haversine distance between session $j$ and either proxy location.

\paragraph{High-frequency donor definition.}
Because invitation burden is penalized only for non-high-frequency donors, we classify donors using a sex-specific annual frequency rule adapted from the STRIDE frequent-donor definition: male donors are considered high-frequency if they made at least 3 donations in the previous 12 months, and female donors if they made at least 2 donations in the previous 12 months \cite{bialkowskiStrategiesReduceIron2015}. 

\paragraph{Objective.}
The implemented BILP minimizes a weighted objective that favors geographically convenient assignments( donor--session distances minimization), discourages repeated invitations to non-high-frequency donors, and penalizes invitations to donors with prior severe adverse reactions:
\begin{equation}
\min \;
W_{\mathrm{dist}}\sum_{(i,j)\in\mathcal{A}} \delta_{ij}x_{ij}
\;+\;
W_{\mathrm{inv}}\sum_{i\in\mathcal{D}_{\mathrm{nonHF}}} y_i
\;+\;
W_{\mathrm{adv}}\sum_{(i,j)\in\mathcal{A}} a_i x_{ij},
\label{eq:main_objective}
\end{equation}
where $a_i\in\{0,1\}$ indicates whether donor $i$ has a recorded severe adverse reaction, $y_i\in\{0,1\}$ is an auxiliary variable that activates when a non-high-frequency donor receives more than one invitation, and $W_{\mathrm{dist}}, W_{\mathrm{inv}}, W_{\mathrm{adv}}$ are non-negative weights. In the experiments reported here we set $W_{\mathrm{dist}}=1.0$, $W_{\mathrm{inv}}=1.0$, and $W_{\mathrm{adv}}=10.0$.

\paragraph{Multiple-invitation penalty for non-high-frequency donors.}
For each non-high-frequency donor $i$ with at least two feasible donor--session pairs, we introduce a binary indicator $y_i$ and enforce
\begin{equation}
\sum_{j:(i,j)\in\mathcal{A}} x_{ij} - 1 \;\le\; (n_i-1)y_i,
\end{equation}
where $n_i = |\{j:(i,j)\in\mathcal{A}\}|$. This makes $y_i=1$ whenever donor $i$ receives two or more invitations, allowing the objective to penalize invitation burden without forbidding it completely.

\paragraph{Constraints.}
Solutions must satisfy the following requirements.

\textbf{(1) Session capacity.}
Each session $j$ has residual capacity $C_j$ expressed in expected-attendance units. We therefore require
\begin{equation}
\sum_{i\in\mathcal{D}:(i,j)\in\mathcal{A}} p_i x_{ij} \le C_j,
\quad \forall j\in\mathcal{S}.
\label{eq:capacity}
\end{equation}

\textbf{(2) Monthly demand by blood type.}
Let $\mathcal{M}$ be the set of months in the planning horizon, $\mathcal{S}_m \subseteq \mathcal{S}$ the sessions in month $m$, and $\mathcal{D}_g \subseteq \mathcal{D}$ the donors of ABO/Rh group $g$.
\paragraph{Demand components and donation-equivalent targets.}
Hospital demand is recorded by blood component. In our study, \textsc{CE} denotes erythrocyte concentrate, including units prepared both with and without buffy coat, and \textsc{CPP} denotes pooled platelet concentrates. These are the main consumed blood components in our data(Supplementary Figure~\ref{fig:supp_components}) and therefore the main demand drivers considered in the planning model. To express monthly demand in whole-blood donation-equivalent units, we assume that one \textsc{CPP} requires five whole-blood donations. Because a single whole-blood donation contributing to a platelet pool can also yield an erythrocyte unit, we do not sum the two component demands. Instead, for each month $m$ and blood group $g$, we define the donation-equivalent demand target as
\begin{equation}
D_{mg} = \max\!\left(d^{\mathrm{CE}}_{mg},\, 5\,d^{\mathrm{CPP}}_{mg}\right).
\label{eq:demand_equivalent}
\end{equation}
Let $R_{mg}$ denote the residual demand target for month $m$ and blood group $g$, expressed in expected-attendance units. The implementation supports two demand modes.

In the \emph{hard-demand} mode, the model attempts to meet a prescribed target and returns infeasible if it cannot be satisfied:
\begin{equation}
\sum_{j\in\mathcal{S}_m}\;\sum_{i\in\mathcal{D}_g:(i,j)\in\mathcal{A}} p_i x_{ij}
\;\ge\;
R_{mg},
\quad \forall m\in\mathcal{M},\; g\in\mathcal{G}.
\label{eq:monthly_demand_hard}
\end{equation}
This mode is useful when operations aim to cover a specific fraction of residual demand with invitations and want infeasibility to signal that the target cannot be attained under the chosen constraints.

In the \emph{soft-demand} mode, non-negative slack variables $s_{mg}$ are introduced so that unmet demand is allowed but penalized:
\begin{equation}
\sum_{j\in\mathcal{S}_m}\;\sum_{i\in\mathcal{D}_g:(i,j)\in\mathcal{A}} p_i x_{ij}
\;+\; s_{mg}
\;\ge\;
R_{mg},
\quad \forall m\in\mathcal{M},\; g\in\mathcal{G},
\label{eq:monthly_demand_soft}
\end{equation}
with $s_{mg}\ge 0$. In this setting, the objective is augmented by
\begin{equation}
W_{\mathrm{dem}}\sum_{m\in\mathcal{M}}\sum_{g\in\mathcal{G}} s_{mg},
\label{eq:slack_penalty}
\end{equation}
where $W_{\mathrm{dem}}$ is a large penalty weight. This formulation encourages the solver to maximize achievable demand fulfillment when full coverage is impossible.

\textbf{(3) Geographic feasibility/convenience (enforced via $\mathcal{A}$).}
For each donor, we evaluate convenience using the two proxy anchor points defined above (last attended brigade and home postal code). A donor--session pair $(i,j)$ is included in $\mathcal{A}$ only if the minimum haversine distance from either proxy location to session $j$ is at most $R=3$ km. Using this radius both enforces a practical convenience criterion and substantially reduces the size of the feasible pair set.

\paragraph{Haversine distance.}
Given donor coordinates $(\phi_1, \lambda_1)$ and session coordinates $(\phi_2, \lambda_2)$ in degrees, the great-circle distance $d$ is computed via the haversine formula:
\begin{align}
    d &= R_\oplus \, c \\
    c &= 2 \arcsin\left(\sqrt{a}\right) \\
    \begin{split}
        a &= \sin^2\left(\frac{\Delta\phi}{2}\right) + \cos\left(\phi_1 \tfrac{\pi}{180}\right)\cos\left(\phi_2 \tfrac{\pi}{180}\right) \\
          &\quad \cdot \sin^2\left(\frac{\Delta\lambda}{2}\right)
    \end{split} \\
    \Delta\phi &= (\phi_2 - \phi_1)\tfrac{\pi}{180}, \quad \Delta\lambda = (\lambda_2 - \lambda_1)\tfrac{\pi}{180}
\end{align}
where $R_\oplus = 6371$ km is the Earth's mean radius.

\textbf{(4) Age eligibility (enforced via $\mathcal{A}$).}
Let $a_i(t_j^{\mathrm{start}})$ denote donor $i$'s age at the start date of session $j$. A pair $(i,j)$ is included in $\mathcal{A}$ only if
\[
18 \le a_i(t_j^{\mathrm{start}}) \le A_i^{\max},
\]
where $A_i^{\max}$ is the donor-specific maximum eligible donation age stored in the registry. Donors under an active medical suspension at the session start date are also excluded.

\paragraph{Session-window representation.}
In the implementation, collection events taking place at the same location within a 14-day period are grouped into a single session-window rather than treated as separate sessions. Each session object therefore stores a start date, an end date, and the set of admissible donation dates within that window, allowing the donor to choose the day on which they wish to attend.

\subsection{Temporal Eligibility Constraints}
\label{sec:temporal}
The most operationally important constraints involve temporal eligibility based on donation history and planned invitations. These rules ensure that invitation plans remain compatible with minimum inter-donation intervals, annual donation limits, and donor-fatigue safeguards. In the exact BILP formulation, they define feasibility conditions for acceptable plans; in the greedy solver, the same rules are checked incrementally during assignment construction.

\textbf{(5) Minimum inter-donation interval.}
Donors must wait at least $\Delta_{\min}=60$ days between consecutive donations. Because sessions are represented as date windows, let $t_j^{\mathrm{start}}$ and $t_j^{\mathrm{end}}$ denote the start and end dates of session $j$, respectively. To ensure feasibility under this representation, if two invited sessions $j_1, j_2$ satisfy
\[
t_{j_2}^{\mathrm{start}} > t_{j_1}^{\mathrm{start}}
\quad\text{and}\quad
t_{j_2}^{\mathrm{start}} - t_{j_1}^{\mathrm{end}} < \Delta_{\min},
\]
we impose
\begin{equation}
x_{i j_1} + x_{i j_2} \le 1.
\label{eq:gap_constraint}
\end{equation}
This prevents assigning a donor to two session-windows whose admissible dates could violate the minimum inter-donation rule.

\textbf{(6) Annual donation limits.}
Within any 365-day window, male donors are limited to 4 donations and female donors to 3. Let $L_i \in \{3,4\}$ be donor $i$'s annual limit, and let $H_i(t)$ be the number of \emph{historical} donations made by donor $i$ in the 365 days preceding date $t$ (inclusive). For each donor $i$ and each session end date $t_j^{\mathrm{end}}$ we require:
\begin{equation}
\sum_{j' \in \mathcal{S}:\; t_j^{\mathrm{end}}-365 < t_{j'}^{\mathrm{end}} \le t_j^{\mathrm{end}}} x_{ij'} \le L_i - H_i(t_j^{\mathrm{end}}).
\label{eq:annual_limit}
\end{equation}

\textbf{(7) Annual invitation cap.}
To mitigate donor fatigue from over-solicitation, we may cap the number of invitations any donor may receive within a rolling 365-day window:
\begin{equation}
\sum_{j:(i,j)\in\mathcal{A}} x_{ij} \le I_i^{\mathrm{rem}},
\quad \forall i \in \mathcal{D},
\label{eq:invite_cap}
\end{equation}
where $I_i^{\mathrm{rem}}$ denotes donor $i$'s remaining invitation budget for the year after accounting for invitations already sent. In the experiments where this mechanism is activated, we use a baseline cap of five invitations per year, informed by the practice of Sanquin, the Dutch national blood service, which operates a voluntary, non-remunerated system structurally comparable to Portugal's IPST and caps personal whole-blood donation invitations at five per year \cite{VeelgesteldeVragenSanquin}. Under this regime, approximately 50\% of personally invited donors attend \cite{weversCharacteristicsDonorsWho2014}, suggesting that a disciplined invitation cap is compatible with high response rates.

\subsection{Penalty for Inviting Donors with Prior Adverse Reactions}
\label{sec:adverse}

Donors with a recorded severe adverse reaction are not excluded from the feasible pool, since doing so could make some instances infeasible. Instead, they are retained as feasible candidates but penalized through the objective via the term
\[
W_{\mathrm{adv}}\sum_{(i,j)\in\mathcal{A}} a_i x_{ij},
\]
with $W_{\mathrm{adv}}=10.0$ in our experiments. This encourages the solver to prefer non-adverse reaction donors whenever feasible, while preserving flexibility when the candidate pool is limited.

\subsection{Greedy Heuristic}
\label{sec:greedy}
Because the exact BILP can become computationally demanding on full-scale instances, we also implement a greedy heuristic that constructs invitation plans incrementally. The heuristic first generates feasible donor--session pairs, groups them by month and blood type, and prioritizes the scarcest demand classes first. Within each class, feasible pairs are considered lexicographically, favoring donors without prior severe adverse reactions and shorter donor--session distances. Each assignment is accepted only if it remains compatible with session capacity, inter-donation gap, and annual donation-limit constraints.

\FloatBarrier
\begin{algorithm}[ht!]
\caption{Scarcity-Driven Greedy Donor--Session Assignment}
\label{alg:greedy}
\small
\begin{algorithmic}
\REQUIRE Donors $\mathcal{D}$, sessions $\mathcal{S}$, residual demand $R_{mg}$ for each month $m$ and blood group $g$
\ENSURE Invitation set $P \subseteq \mathcal{D} \times \mathcal{S}$

\STATE $P \gets \emptyset$

\STATE \textbf{Phase 1: Feasible pair generation}
\FOR{each session $j \in \mathcal{S}$}
    \FOR{each donor $i \in \mathcal{D}$}
        \IF{$(i,j)$ satisfies age, suspension, distance, and last-donation-gap constraints}
            \STATE add pair $(i,j)$ to feasible set $\mathcal{F}$
            \STATE assign $(i,j)$ to demand class $(m(j), g(i))$
        \ENDIF
    \ENDFOR
\ENDFOR

\STATE \textbf{Phase 2: Demand prioritization}
\FOR{each demand class $(m,g)$ with $R_{mg} > 0$}
    \STATE compute scarcity score $\sigma_{m,g} \gets |\mathcal{F}_{m,g}| / R_{mg}$
\ENDFOR
\STATE sort demand classes in increasing order of $\sigma_{m,g}$

\STATE \textbf{Phase 3: Greedy assignment}
\FOR{each demand class $(m,g)$ in sorted order}
    \STATE let $\mathcal{C}_{m,g}$ be the feasible pairs in class $(m,g)$
    \STATE sort $\mathcal{C}_{m,g}$ lexicographically by
    \STATE \quad (i) non-adverse reaction donors first
    \STATE \quad (ii) shorter travel distance second
    \FOR{each pair $(i,j) \in \mathcal{C}_{m,g}$ in that order}
        \IF{demand for $(m,g)$ is already satisfied}
            \STATE \textbf{break}
        \ENDIF
        \IF{$(i,j)$ remains feasible with respect to session capacity, inter-donation gap, annual donation limits and the annual invitation cap}
            \STATE add $(i,j)$ to $P$
            \STATE update fulfilled demand for $(m,g)$
            \STATE update donor $i$'s assignment history
            \STATE update session $j$'s used capacity
        \ENDIF
    \ENDFOR
\ENDFOR

\STATE \textbf{return} $P$
\end{algorithmic}
\end{algorithm}
\FloatBarrier

\paragraph{Implementation notes.}
Demand and capacity are tracked in probability-weighted units: each donor $i$ contributes its attendance probability $p_i \in (0,1]$ toward both the residual demand target $R_{mg}$ and the session capacity $C_j$, rather than counting invitations as unit contributions. The annual-limit check evaluates rolling 365-day windows over the union of historical donation dates and all dates already present in $P$ for donor $i$. The inter-donation-gap check requires that any newly assigned session $j$ be at least $\Delta_{\min}=60$ days away from all prior historical and planned donations of donor $i$. The feasibility check also enforces the donor-specific remaining invitation budget.

\subsection{Prospective Invitation Pipeline}
\label{sec:prospective_pipeline}

The retrospective analysis in Section~\ref{sec:results} uses ex-post demand. Operational deployment requires a fully prospective pipeline that estimates how many invitations to send without knowing future demand or organic attendance. The pipeline proceeds in three steps:

\paragraph{Step 1 – Demand targeting.} For each month $m$ and blood group $g$, the demand target $D_{mg}$ is estimated from the historical distribution using a trend-corrected quantile at level $\alpha \in (0,1)$, as described in Section~\ref{sec:results}. Higher $\alpha$ yields more conservative (higher) targets, providing a buffer against forecast uncertainty.

\paragraph{Step 2 – Organic supply subtraction.} The expected number of donors who will attend without being invited is estimated by the organic attendance pipeline: the MLP-based returning-donor model and the Holt--Winters first-time-donor forecast (Section~\ref{sec:organic_attendance}). Their blood-type-disaggregated expected contributions are subtracted from $D_{mg}$ to obtain the \emph{residual demand} $R_{mg} = \max(0,\, D_{mg} - \hat{O}_{mg})$, where $\hat{O}_{mg}$ is the predicted organic supply. Residual session capacity $\tilde{C}_j$ is similarly computed by subtracting the expected organic attendance at each site from the nominal capacity $C_j$.

\paragraph{Step 3 – Invitation planning.} The chosen strategy (Gurobi or Greedy) is run on the residual problem: donors predicted to attend organically are excluded from the invitation pool, and the solver fills the residual demand $R_{mg}$ subject to residual capacity $\tilde{C}_j$. Show-up probability can be  parameterised by donor type (high-frequency vs.\ lapsed).

\subsection{Implementation Details}
\label{sec:impl}

\paragraph{Solver configuration and compute.}
We solve the resulting \emph{binary integer linear program} with Gurobi Optimizer version 13.0.1 build v13.0.1rc0 \cite{gurobi}. In the reported experiments, Gurobi was run with \texttt{Threads=4} and \texttt{MIPFocus=2}, prioritizing the search for high-quality feasible solutions. All experiments were run on a Runpod machine with an AMD EPYC 7763 and 250\,GB RAM.

\paragraph{AI-assisted geocoding.}
Since CP7 codes for collection sites are not available in the registry, GPT-5.2 agents were used to leverage the available collection-site descriptors and identify the most probable address/location and its corresponding CP7 postal code. The resulting CP7 entries were then mapped to latitude--longitude coordinates using the Eurostat GISCO postal code point dataset \cite{PostalCodesGISCO}, enabling distance computation via the haversine formula.

\paragraph{Rolling-window planning.}
We adopt a rolling-window strategy, deciding invitations in sequential 4-month blocks (e.g., Jan--Apr, May--Aug, Sep--Dec 2020). In the retrospective experiments, donor histories are updated using the donations scheduled within each solved window as a simulation proxy, so that subsequent windows correctly account for newly assigned donations when evaluating eligibility. In practical deployment, these updates should instead rely on realized attendance and observed donation outcomes.

\section{Proposed Operational Framework}
\label{sec:operational_framework}

Based on the computational findings, we outline an operational pipeline for integrating the proposed decision support tool into IPST planning workflows. The pipeline requires a monthly demand target as input, which in our framework is derived from historical demand using blood-type-specific target setting informed by past consumption patterns. In practical settings where this component is unavailable, a simple fallback baseline is to use observed consumption from the corresponding period of the previous year, optionally adjusted by a scaling factor.

\paragraph{1. Data Infrastructure Requirements.} To support the decisions, the blood institute must maintain a consolidated ``contactable'' database of donors that opted-in. Crucially, this database requires: \begin{itemize} \item \textbf{Geocoding:} Latitude and longitude of atleast one preferred location indicated by the donor (e.g., donor’s home or workplace). This will be used to invite the donor to convenient close locations. \item \textbf{Status Tracking:} Periodic  updates on donor suspensions and donation history. \end{itemize}

\paragraph{2. Rolling-Window Execution.}
Because both donor eligibility and expected demand evolve over time, the framework is executed in sequential planning windows rather than over the full year at once. A 4-month rolling horizon is recommended.
\begin{enumerate}
    \item \textbf{Initialization:} At the start of each window (e.g., Jan 1st), load the current donor state, including donation history, suspensions and adverse-reaction records.

    \item \textbf{ML Forecasting:} Before optimization, estimate the expected \emph{organic} supply for the window. This includes: (i) donor-level prediction of returning-donor attendance probabilities using the trained ML model; (ii) forecasting the inflow of first-time donors; and (iii) estimating how predicted organic attendance is distributed across collection sites in order to derive residual site capacities. These components are then aggregated by month and blood group to estimate the  expected supply for the window from donors who are likely to attend without invitation.

    \item \textbf{Residual target construction and parametrization:} Operations managers define the window length (max.\ 4 months), the demand coverage $\rho$, and a geographic cutoff radius $R$. Forecasted organic supply is subtracted from the target demand to obtain the residual demand to be covered by invitations, and predicted organic attendance is subtracted from nominal session capacity to obtain residual capacity. Alternatively, when full residual-demand coverage is unlikely, the soft-demand formulation can be used so that the optimization instead maximizes achievable fulfillment under the imposed constraints. To balance convenience and feasibility, we recommend a short radius-sweep: solve first with a conservative radius (e.g., $R=3$ km); if infeasible, increase $R$ in a small set of steps (e.g., 4--6 km) until feasibility is achieved, then adopt the smallest feasible $R$. This preserves donor convenience while controlling the size of the feasible pair set $\mathcal{A}$.

    \item \textbf{Solving:} The solver generates an optimized invite list for the residual problem in the current window, or returns that the instance is infeasible under the chosen parameters.

    \item \textbf{Execution and update:} Invitations are sent. As the window progresses, the database must be updated with \textit{actual} attendance. Because the planning stage relies on predicted organic attendance and expected invited show-up, discrepancies between predicted and realized behavior will occur. Donors who decline or fail to attend may become eligible sooner than anticipated, while realized organic attendance may differ from forecasts. Therefore, the computation for the \textit{subsequent} window (e.g., May--Aug) must be re-run using updated ground-truth histories and newly observed attendance outcomes.
\end{enumerate}

\section{Results}
\label{sec:results}
IPST’s blood-collection activity is organized through three main regional Centers of Blood and Transplantation (CST), commonly referenced in operational reporting as Lisbon (SL), Porto (SP), and Coimbra (SC).
We evaluate the proposed decision support framework using historical data from the Lisbon (SL) region for the year 2020. The evaluation focuses on three dimensions: (1) the structural supply-demand imbalance in the region, (2) the computational tractability of the BILP formulation, and (3) the potential operational gains from strategic donor selection.

\subsection{Supply-Demand Imbalance and Donor Potential}
\begin{figure}[ht]
    \centering
    \includegraphics[width=\linewidth]{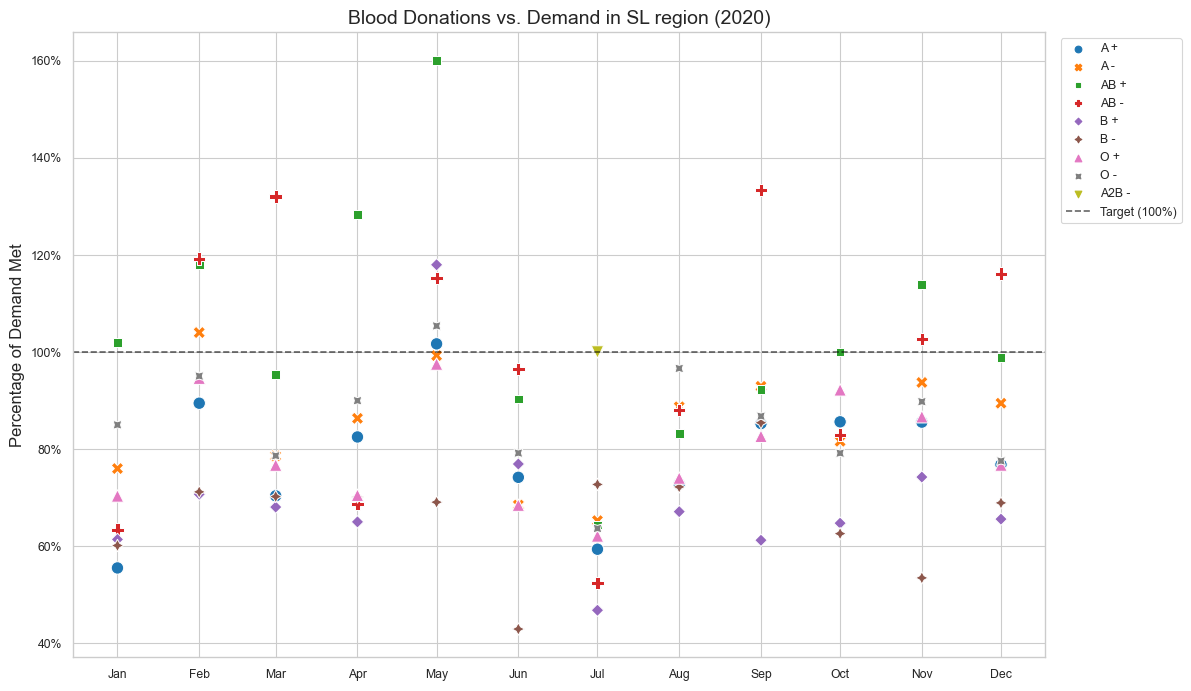}
    \caption{Blood donations vs. Demand in the SL region (2020). \normalfont The dashed line represents 100\% of demand met. Consumption of units of several blood types consistently exceed donations.}
    \label{fig:supply_demand}
\end{figure}

At the national level, if we aggregate demand and donations, Portugal was able to satisfy nearly all blood demand in 2020 (Supplementary Figure~S1). However, this national self-sufficiency masks operational regional heterogeneity. While some operational zones are able to remain broadly self-sufficient, the Lisbon region is historically characterized by a structural deficit in blood collection and often depends on inter-regional transfers to meet hospital demand \cite{portugalReservasSangueEstao2024}. Lisbon hosts a high concentration of centers that perform procedures requiring massive amounts of blood products and that serve as end-of-line hospitals, handling some of the country's most complex clinical cases\cite{CHLC_PAO_2017, CHLN_RelatorioContas_2017}. Figure~\ref{fig:supply_demand} illustrates the monthly  supply versus demand for 2020.  Actual collections consistently fell below 100\% of the regional need.

\begin{figure}[ht]
    \centering
    \includegraphics[width=\linewidth]{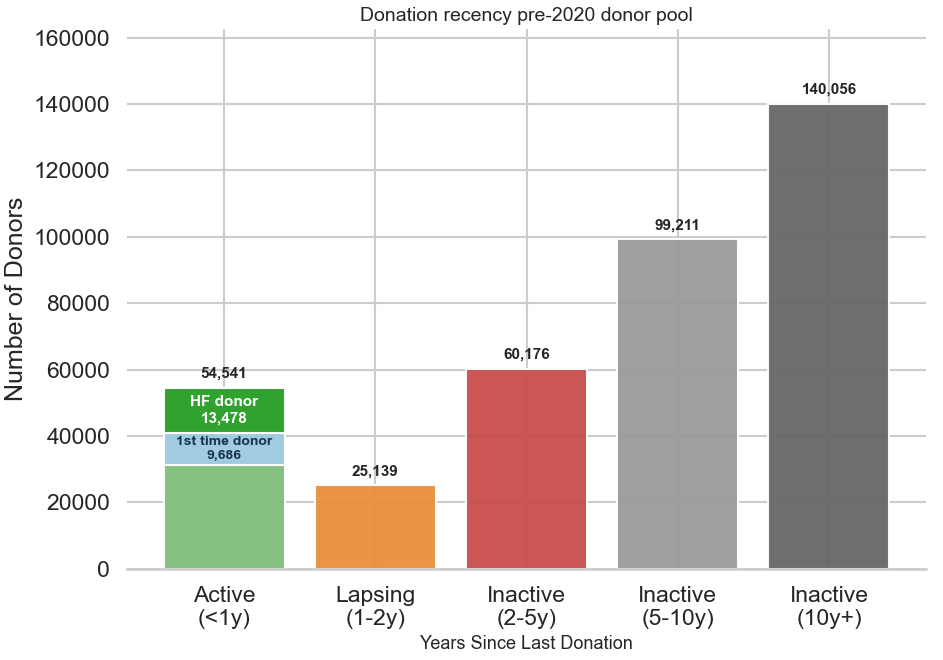}
    \caption{Distribution of last-donation recency \normalfont prior to January 2020, restricted to donors who were age-eligible to donate in 2020. Donors are grouped into active, lapsing and inactive categories based on recency of their last donation; HF denotes high-frequency donors. A substantial portion of the pool had not donated for more than 10 years, indicating considerable theoretical reactivation potential.}
    \label{fig:donor_recency}
\end{figure}

Figure~\ref{fig:donor_recency} organizes the donor pool by recency of last donation. 
We follow the donor-status nomenclature used in the DOMAINE donor-management framework \cite{domainemanual2}, where \emph{active} donors donated within the last 12 months, \emph{lapsing} donors donated within 
the last 24 months but not within the last 12 months, and \emph{inactive} donors had no donation in 
the last 24 months. A vast majority of the pool consists of ``Inactive'' donors who have not donated in over 2 years, including over 140,000 donors who have been inactive for more than 10 years but remained within the eligible age range in 2020.

\begin{figure}[ht]
    \centering
    \includegraphics[width=\linewidth]{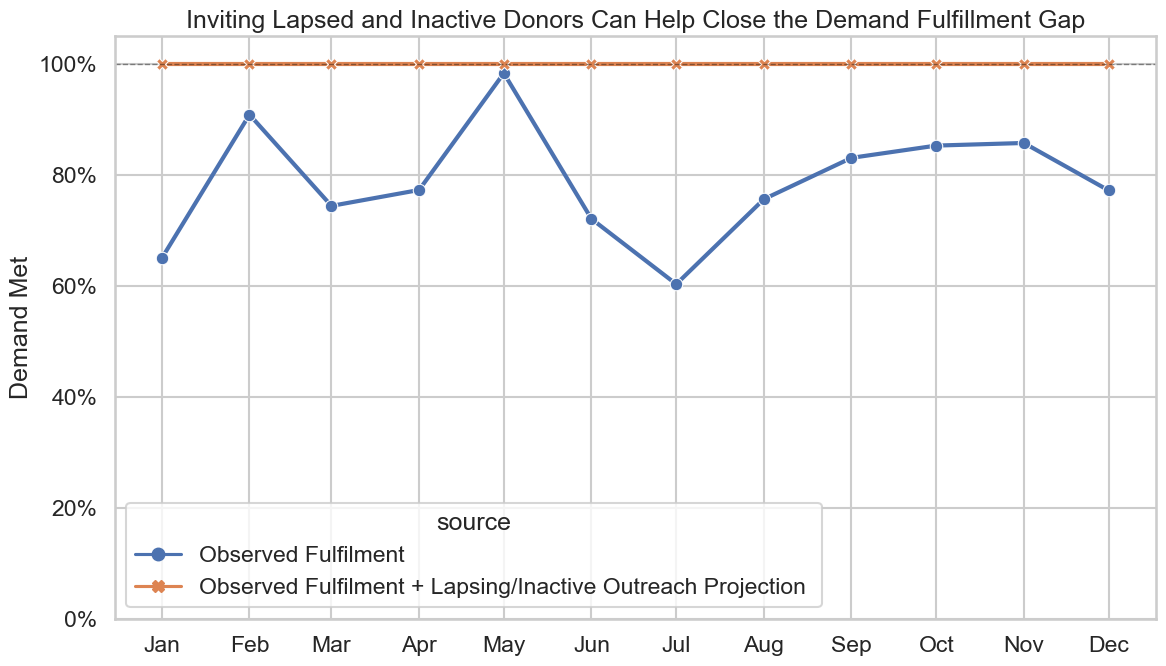}
    \caption{Potential vs.\ observed demand fulfillment in the SL region (2020). \normalfont The red line shows observed monthly fulfillment based on actual collections, while the green dashed line shows the simulated BILP result after targeted outreach to eligible inactive/lapsing donors. Under the assumed 5\% attendance probability, the donor pool appears sufficient to close the regional supply gap.}
    \label{fig:potential_fulfillment}
\end{figure}

To evaluate the operational potential of this underutilized pool, we simulated a targeted outreach campaign directed exclusively at lapsed and inactive donors in the SL operational region for 2020. In this scenario, observed historical donations were fixed as a baseline, and the Gurobi optimization pipeline was tasked with fulfilling the remaining unmet monthly demand. The solver specifically targeted eligible inactive donors (those with no recorded donations in 2020) residing within a 3~km geographic radius of the collection sites. Assuming a conservative 5\% attendance probability for these re-engaged donors---noting that reported response rates to personal invitations in Europe have ranged from 5\% to 80\% \cite{weversCharacteristicsDonorsWho2014}---Figure~\ref{fig:potential_fulfillment} compares the observed baseline fulfillment against the projected total fulfillment. The results demonstrate that complementing actual attendance with constraint-aware, targeted outreach to inactive donors successfully closes the supply gap, meeting 100\% of the demand targets across the year.

While these results highlight the substantial theoretical capacity of the inactive donor pool, it is important to note that this simulation is fundamentally retrospective. The solver had access to the exact monthly demand targets ex-post. For this decision support framework to be deployed in a forward-looking operational setting—where the goal is to proactively fulfill unmet demand—it must be paired with robust demand forecasting. To operationalise this approach, blood centers must set future demand targets based on historical consumption trends and reliably estimate the expected show-up rates of the targeted demographic to ensure sufficient invitations are scheduled.

\subsection{Demand Forecasting}

To move from the retrospective complement experiment to a prospective planning setting, future monthly demand targets must be estimated without access to ex-post consumption. We therefore evaluated monthly demand predictability on the 2007--2024 Lisbon demand panel, disaggregated by ABO/Rh group and by the two planning-relevant product families (\textsc{CE} and \textsc{CPP}). For each evaluation year $y \in \{2009,\dots,2024\}$, all predictive quantities were estimated using only years $<y$, so that each year was assessed in a strict expanding-window setting.

\begin{figure}[!htbp]
    \centering
    \includegraphics[width=\linewidth]{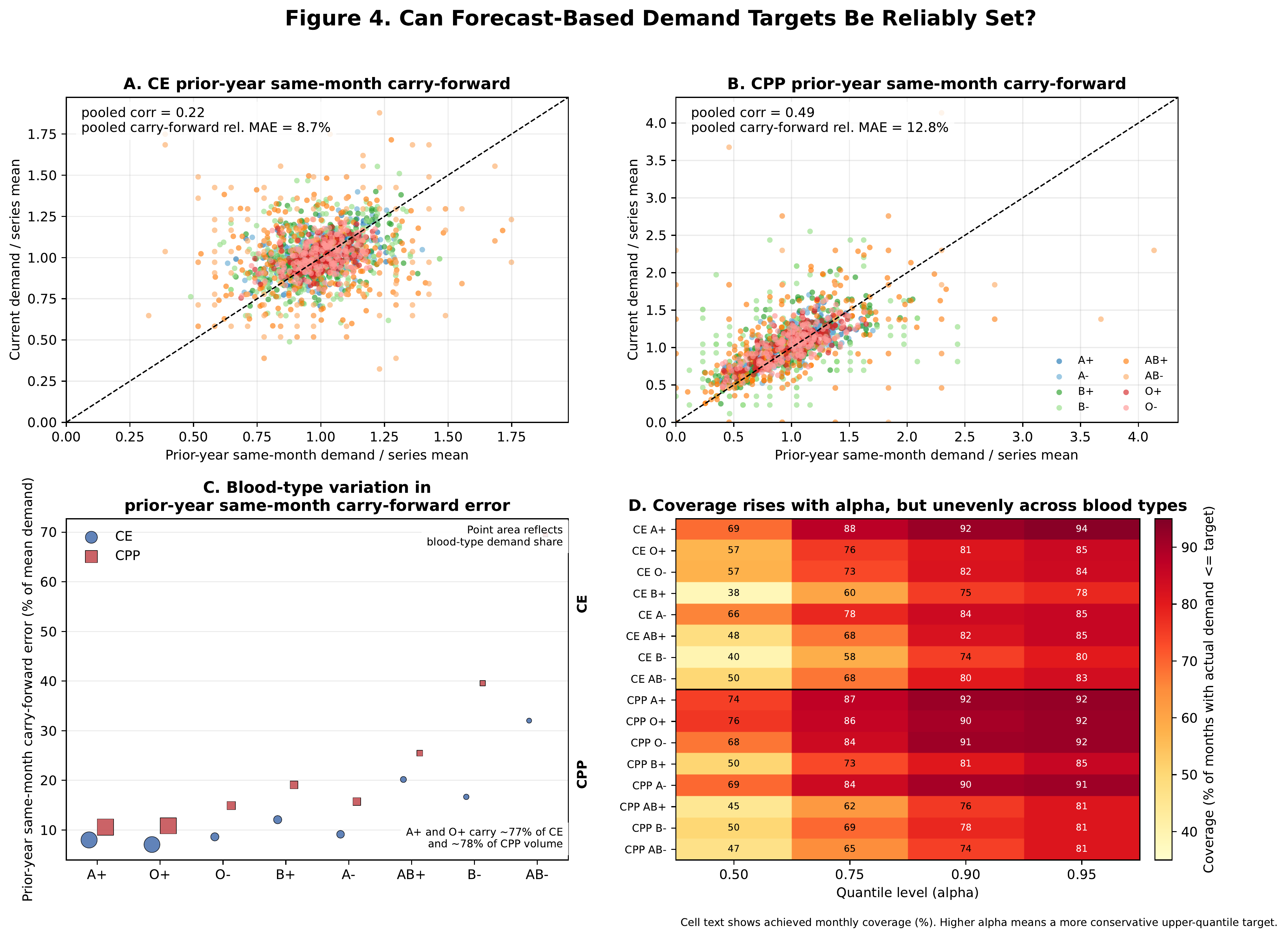}
    \caption{Forecast-based demand targets. \normalfont Panels A-B show pooled prior-year same-month carry-forward performance for \textsc{CE} and \textsc{CPP} forecasting. Panel C shows that carry-forward forecasting error is lowest in the high-volume blood groups and much higher in rare groups. Panel D shows achieved monthly coverage for trend-corrected quantile targets across blood types and quantile levels $\alpha$: higher $\alpha$ gives a more conservative target, but the coverage gains are uneven across blood types.}
    \label{fig:demand_forecast_summary}
\end{figure}

Figure~\ref{fig:demand_forecast_summary} asks whether forecast-based demand targets can be set reliably. Panels A and B use the prior-year same-month carry-forward rule as a transparent baseline. At the pooled level, this rule still captures some month-to-month structure, but it is far from uniformly reliable: the pooled normalized correlation is $r=0.22$ for \textsc{CE} and $r=0.49$ for \textsc{CPP}, with pooled carry-forward relative MAE of 8.7\% and 12.8\%, respectively. Panel C shows why pooled summaries are not enough. Carry-forward error is materially lower in the high-volume groups A+ and O+ than in the rare groups; for example, prior-year same-month carry-forward relative MAE is 8.0\% for A+ \textsc{CE}, 7.1\% for O+ \textsc{CE}, 10.6\% for A+ \textsc{CPP}, and 10.8\% for O+ \textsc{CPP}, but rises to 32.0\% for AB$-$ \textsc{CE} and 69.6\% for AB$-$ \textsc{CPP}. The main implication is that demand predictability is heterogeneous across blood types and should not be summarized by a single pooled rule alone.

Because shortage cost is asymmetric, the planning problem is not to choose one point forecast and stop there, but to convert historical demand into conservative blood-type-specific targets. We do this using month-specific quantiles with selective linear trend correction. The quantile level $\alpha$ controls how conservative the target is: higher $\alpha$ increases protection against shortages, but also raises over-targeting. Panel D shows that the coverage gain from increasing $\alpha$ is real but uneven across blood types, so a single universal level is not appropriate.

\subsection{Organic Attendance Forecasting}
\label{sec:organic_attendance}

\begin{figure}[!t]
    \centering
    \includegraphics[width=\linewidth]{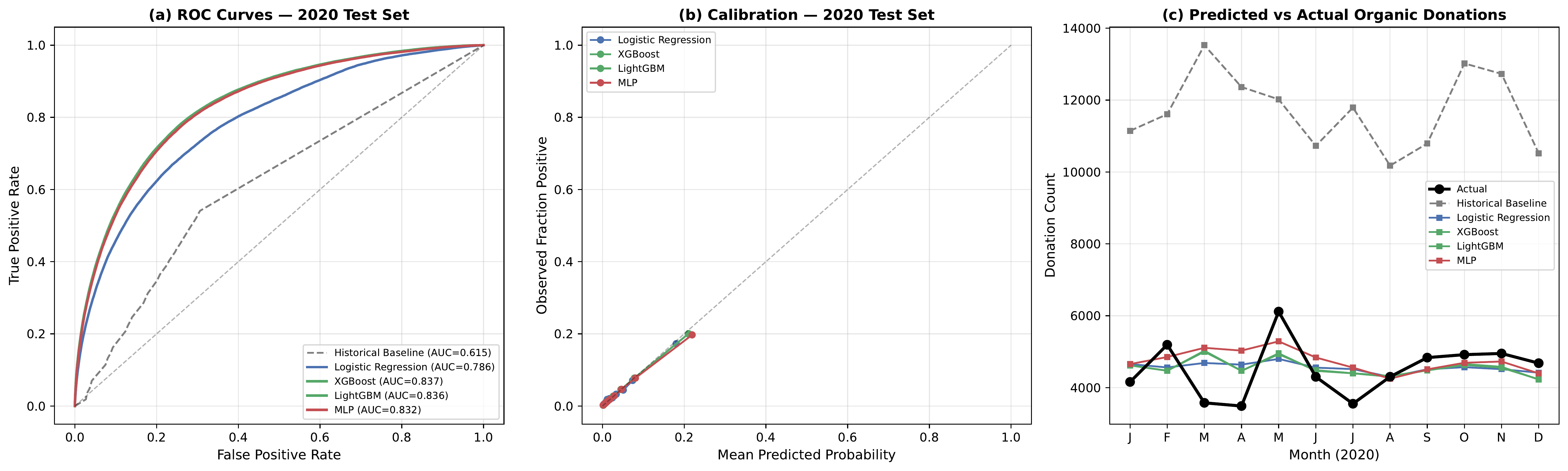}
    \caption{Organic attendance forecasting for the SL operational region, \normalfont evaluated on the 2020 donor-month test set after selecting an expanding training window from pre-2020 backtests and retuning the tree ensembles without class-balanced weighting. \bfseries (a) ROC curves \normalfont show that logistic regression, MLP, XGBoost, and LightGBM all discriminate substantially better than the historical baseline. \bfseries (b) Calibration curves \normalfont show that, after removing the probability distortion induced by class-balanced tree weighting, the three ML models are all reasonably calibrated, with retuned XGBoost and LightGBM slightly outperforming the MLP on Brier score. \bfseries (c) Monthly aggregate organic attendance forecasts \normalfont obtained by summing donor-level probabilities. The MLP remains best on monthly count accuracy (MAE 720 donors, 13.6\% relative MAE), while retuned XGBoost (MAE 741, ROC-AUC 0.822, PR-AUC 0.242, Brier 0.0369) and retuned LightGBM (MAE 745, ROC-AUC 0.822, PR-AUC 0.241, Brier 0.0370) are very close.}
    \label{fig:organic_attendance}
\end{figure}

Prospective invitation planning also requires an estimate of how many eligible donors will donate without being contacted. We therefore framed organic attendance as a donor-month prediction problem on the SL operational region, using monthly windows over the 2007--2020 donor history. For each eligible donor and target month, features were computed strictly from pre-window history and included recency of last donation, total donation count, 1-year and 2-year donation frequency, eligibility rate, mean and standard deviation of inter-donation gaps, prior adverse-reaction history, same-calendar-month history, years active, age, and sex. The primary target for Figure~\ref{fig:organic_attendance} is donation \emph{attendance} rather than accepted collections, because from an operational standpoint the invitation system first needs to estimate who will present to donate; completion rates can then be handled as a second layer.

We first compared expanding, rolling 3-year, and rolling 5-year training windows on pre-2020 backtests; the full sweep summary is reported in Supplementary Table~\ref{tab:supp_organic_backtest}. The preferred window was not identical for every model: expanding was best for XGBoost and LightGBM, rolling 5-year was marginally best for logistic regression, and the historical baseline was invariant. In the implemented workflow, XGBoost served as the primary strategy selector for this screening step, so the expanding window was retained for the final 2020 figure; the MLP entered in the subsequent tuning stage after the window policy had already been fixed. Hyperparameters were then tuned on the 2019 validation year and the selected models were refit on all donor-month rows from 2007--2019 before evaluation on 2020. 

The three ML families become genuinely competitive rather than qualitatively different (Figure~\ref{fig:organic_attendance}). Retuned unweighted XGBoost achieved the strongest donor-level discrimination on the 2020 test set (ROC-AUC 0.822, PR-AUC 0.242) and the best overall Brier score (0.0369), with retuned unweighted LightGBM very close behind (ROC-AUC 0.822, PR-AUC 0.241, Brier 0.0370). The MLP remained the most accurate model at the monthly aggregate level, delivering the lowest monthly attendance MAE (720 donors; relative MAE 13.6\%), compared with 741 for retuned XGBoost and 745 for retuned LightGBM. Logistic regression was still competitive but slightly weaker on both ranking and aggregate error (monthly MAE 760 donors; Brier 0.039). The historical same-month baseline was clearly inadequate by comparison (ROC-AUC 0.612; monthly MAE 8{,}237 donors).

The key implication is that the model used to feed the downstream invitation optimizer should be selected on a joint criterion combining donor-level ranking quality, calibration, and monthly aggregate count accuracy, rather than ROC-AUC alone. Under that criterion, the margin between the leading models is narrow: XGBoost is slightly stronger on donor-level ranking and Brier score, whereas the MLP is slightly stronger on aggregate monthly count accuracy. Because the downstream pipeline consumes summed donor probabilities as expected monthly organic supply, we retain the MLP as the primary operational model.

\paragraph{First-time donor inflow.}
The donor-month return model above only scores donors who have at least one prior donation in the SL system; individuals whose very first donation falls in the target month are structurally excluded. To quantify this missing component, we extracted the monthly count of first-time donors.

First-time donors contributed approximately 640 per month during the three pre-evaluation years (2017--2019), representing $\sim$8--10\% of total monthly organic attendance---a non-negligible supply term that, if omitted, would cause the downstream optimizer to systematically over-invite. At the annual level (Supplementary Figure~\ref{fig:supp_annual_first_time}), the absolute count of first-time donors declined from a peak of $\sim$11{,}500 in 2010 and 2012 to $\sim$7{,}400--8{,}100 in the 2017--2019 period, following an overall downward trend of approximately $-144$ donors per year. However, as a share of total donations, first-time donors have remained relatively stable at 8--10\%, suggesting that overall collection activity has contracted in parallel. The inflow also exhibits within-year seasonality (peaks in March--April and November, troughs in August--September), consistent with institutional recruitment campaigns and university-based blood drives.

\begin{figure}[!htbp]
    \centering
    \includegraphics[width=\linewidth]{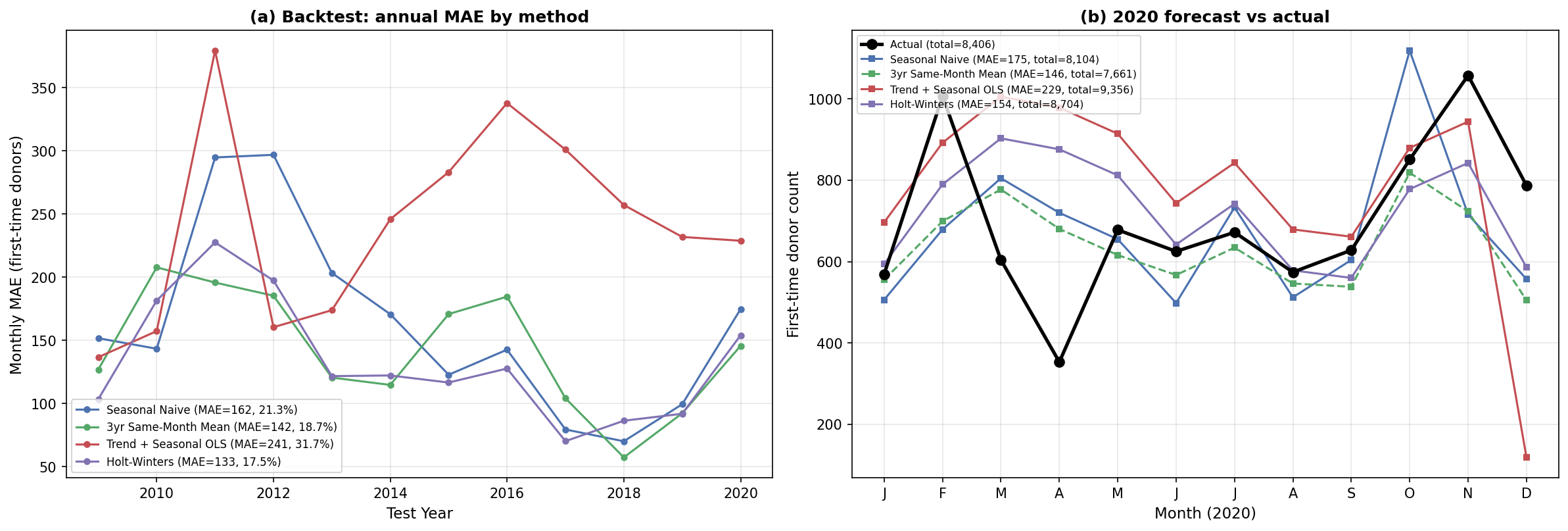}
    \caption{Forecasting first-time donor inflow for the SL operational region. (a) Leave-one-year-out backtest MAE across 2009--2020 for four forecasting methods; \normalfont legend reports overall MAE and relative MAE. Holt--Winters (MAE\,=\,133, 17.5\%) is the most accurate across the full backtest. \bfseries (b) Prospective 2020 forecasts versus actual monthly counts. \normalfont The COVID-19 disruption in March--April is the dominant source of forecast error across all methods. Legend reports monthly MAE and annual total for each method against the actual total of 8{,}406.}
    \label{fig:first_time_forecast}
\end{figure}

To forecast monthly first-time donor counts in a prospective setting, we compared four methods in a leave-one-year-out expanding-window backtest on 2009--2020 (Figure~\ref{fig:first_time_forecast}): Holt--Winters additive forecasting achieved the lowest aggregate MAE (133 donors/month, 17.5\% relative MAE), followed by the 3-year same-month mean (142, 18.7\%), seasonal naive (162, 21.3\%), and OLS trend + seasonal dummies (241, 31.7\%). For the prospective 2020 evaluation, the 3-year same-month mean delivered the best monthly MAE (146), with an annual total of 7{,}661 versus 8{,}406 actual---an under-prediction of 8.9\%, largely attributable to the COVID-19 disruption.

The blood-type composition of first-time donors is estimated from the three-year trailing proportions (A+ 37.9\%, O+ 35.0\%, B+ 7.9\%, O$-$ 7.5\%, A$-$ 6.2\%, AB+ 3.6\%, B$-$ 1.4\%, AB$-$ 0.5\%). In the operational pipeline, total organic supply for each month and blood type is therefore computed as the sum of (i) the MLP-based returning-donor probability aggregates and (ii) the Holt--Winters first-time-donor forecast allocated by blood-type share.

\paragraph{Session assignment.}
The downstream optimizer requires two complementary inputs: (i)~\emph{site-level capacity estimates}---how many organic donors to expect at each collection site, so that residual capacity available for invited donors can be computed---and (ii)~a \emph{per-donor exclusion list}---which specific donors are predicted to attend organically, so they can be removed from the invitation pool to prevent double-counting. We address each with a dedicated model.

\textbf{Site-level capacity model.} To predict how many organic donors will attend each site in a given month, we train a gradient-boosted regression model (GBR; 200 trees, depth\,=\,4) on 2018--2019 site-month observations. Features include the predicted monthly aggregate of organic donors (from the MLP and Holt--Winters forecasts above), each site's historical share and average attendance (expanding mean over pre-test years), site geographic coordinates (latitude, longitude), and month-of-year. The model is evaluated on 2020 data, where the predicted monthly aggregate replaces the actual total. Feature importance is dominated by the historical share (48\%) and historical average attendance (47\%), with longitude providing a modest geographic correction (5\%); the model effectively learns site-specific baselines modulated by the overall predicted supply.

\textbf{Per-donor assignment for optimizer exclusion.} Each donor in the MLP prediction set is assigned to their historical preferred site---defined as the mode of \texttt{site\_id} across all pre-2020 SL donations. Because many historical sites are no longer active in the evaluation year (2{,}220 of 2{,}570 unique preferred sites), donors assigned to such ``ghost sites'' are remapped to the nearest active site using a tiered geocoding strategy: (i)~the ghost site's postal code is geocoded via a national postcode coordinate table (covering 1{,}876 sites), (ii)~for the remainder, median donor coordinates from the donor registry serve as a proxy (319~sites), and (iii)~the 25~sites with no coordinate source are distributed proportionally. The median remapping distance is 0.5\,km, and 89\% of remappings stay within the same municipality (GADM level-2 spatial join). For each month, the expected returning-donor count at a site is obtained by summing the MLP-predicted probabilities of all donors assigned to that site (expected-value weighting). First-time donors, having no donation history, are distributed proportionally across sites using historical first-time site shares from 2017--2019.

\begin{figure}[!htbp]
    \centering
    \includegraphics[width=\linewidth]{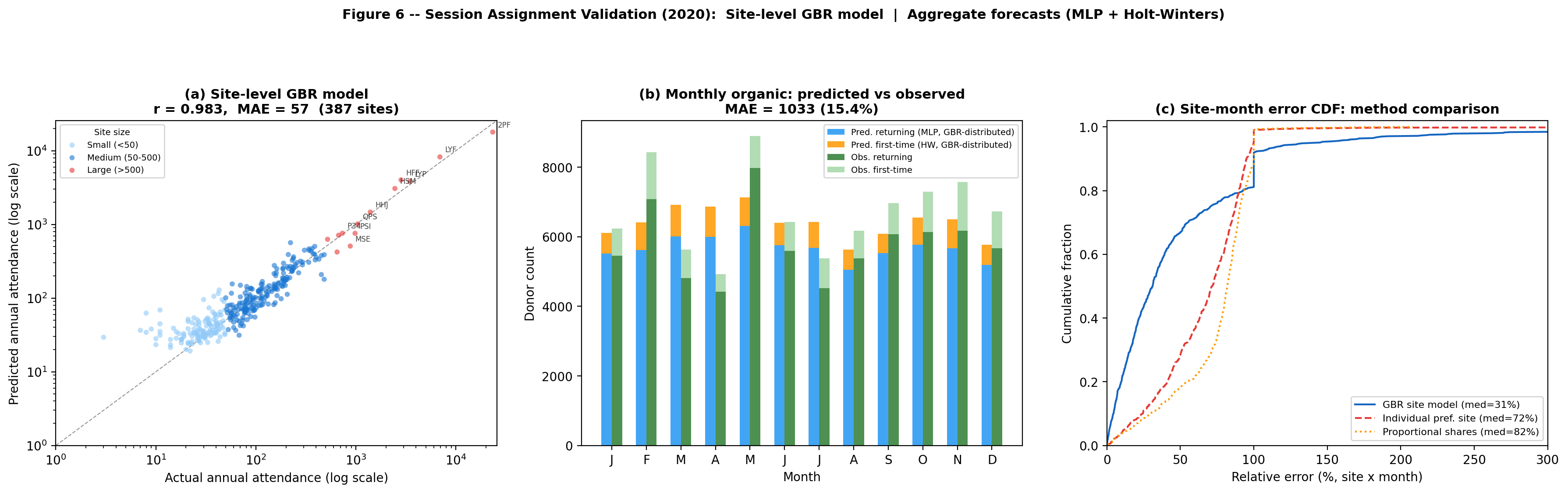}
    \caption{Session assignment validation for the SL operational region, 2020. (a)~Site-level GBR model (with top-down reconciliation): \normalfont predicted versus actual annual attendance per site on log--log axes ($r = 0.983$, MAE\,=\,57, 387 sites); dots are coloured by annual site size. \bfseries (b)~Monthly organic supply: \normalfont predicted returning donors (MLP aggregate, blue) and first-time donors (Holt--Winters forecast, orange) versus observed; monthly MAE\,=\,1{,}033 (15.4\%). The GBR determines how each monthly total is distributed across sites but is invisible at the national level shown here. \bfseries (c)~Cumulative distribution function (CDF) of site$\times$month relative errors comparing three methods: \normalfont GBR site model (median 31\%), individual preferred-site assignment (median 72\%), and proportional historical shares (median 82\%).}
    \label{fig:session_assignment}
\end{figure}

Figure~\ref{fig:session_assignment} validates the site-level GBR model against actual 2020 attendance. To ensure that site-level predictions remain consistent with the more accurate national-level forecasts, the GBR site predictions are rescaled each month so that they sum exactly to the MLP\,+\,Holt--Winters aggregate total---a standard top-down hierarchical reconciliation step \citep{hyndmanOptimalCombinationForecasts2011}. This preserves the GBR's learned site distribution while anchoring totals to the aggregate forecasts. Across 387 active sites, the rescaled GBR achieves a Pearson correlation of $r=0.983$ and a site-level mean absolute error of 57 donors/year---substantially more accurate than the individual preferred-site assignment ($r=0.978$, MAE\,=\,89) or proportional allocation ($r=0.971$, MAE\,=\,80) (Supplementary Table~\ref{tab:session_methods}). Crucially, after reconciliation the GBR also matches the proportional method on monthly aggregate accuracy (MAE\,=\,1{,}033), so no accuracy is sacrificed at any level. The improvement is most pronounced at the site$\times$month level: the GBR's median relative error is 31\%, compared with 72\% for the individual assignment and 82\% for proportional shares (Figure~\ref{fig:session_assignment}c). The dominant remaining error source is SL\_2PF (the fixed Lisbon blood center), which absorbed 29\% of actual 2020 attendance versus 18\% predicted, reflecting a COVID-era shift from cancelled mobile drives to the permanent center that no historical model can capture. Per-donor assignment, while less accurate at the site level, serves a distinct operational purpose: it provides the optimizer with a donor-specific exclusion list so that predicted organic donors are not re-invited.

\subsection{Algorithm Comparison}
\label{sec:solver_comparison}

The BILP formulation provides optimality guarantees but carries non-trivial computational costs that may limit operational deployment on resource-constrained systems. To evaluate whether the greedy heuristic offers a viable substitute, we conducted a controlled comparison of two solvers on the full 2020 Lisbon donor pool, adopting a retrospective residual-demand scenario similar to the potential-fulfillment analysis. Specifically, observed historical donations were fixed as a baseline, and the solvers were tasked with fulfilling the remaining unmet demand without inviting donors who organically donated in 2020. To stress-test the algorithms and make differences in demand fulfillment more evident, we artificially scaled the demand by a factor of 2 ($2\times$). Both strategies operate on identical input data:

\begin{itemize}
  \item \textbf{Greedy}: the lexicographic greedy heuristic Section~\ref{sec:greedy}.
  \item \textbf{Gurobi} (ExplicitAvoidAdverse): the full BILP with adverse-reaction penalty (Section~\ref{sec:adverse}).
\end{itemize}

\begin{figure}[!htbp]
    \centering
    \includegraphics[width=\linewidth]{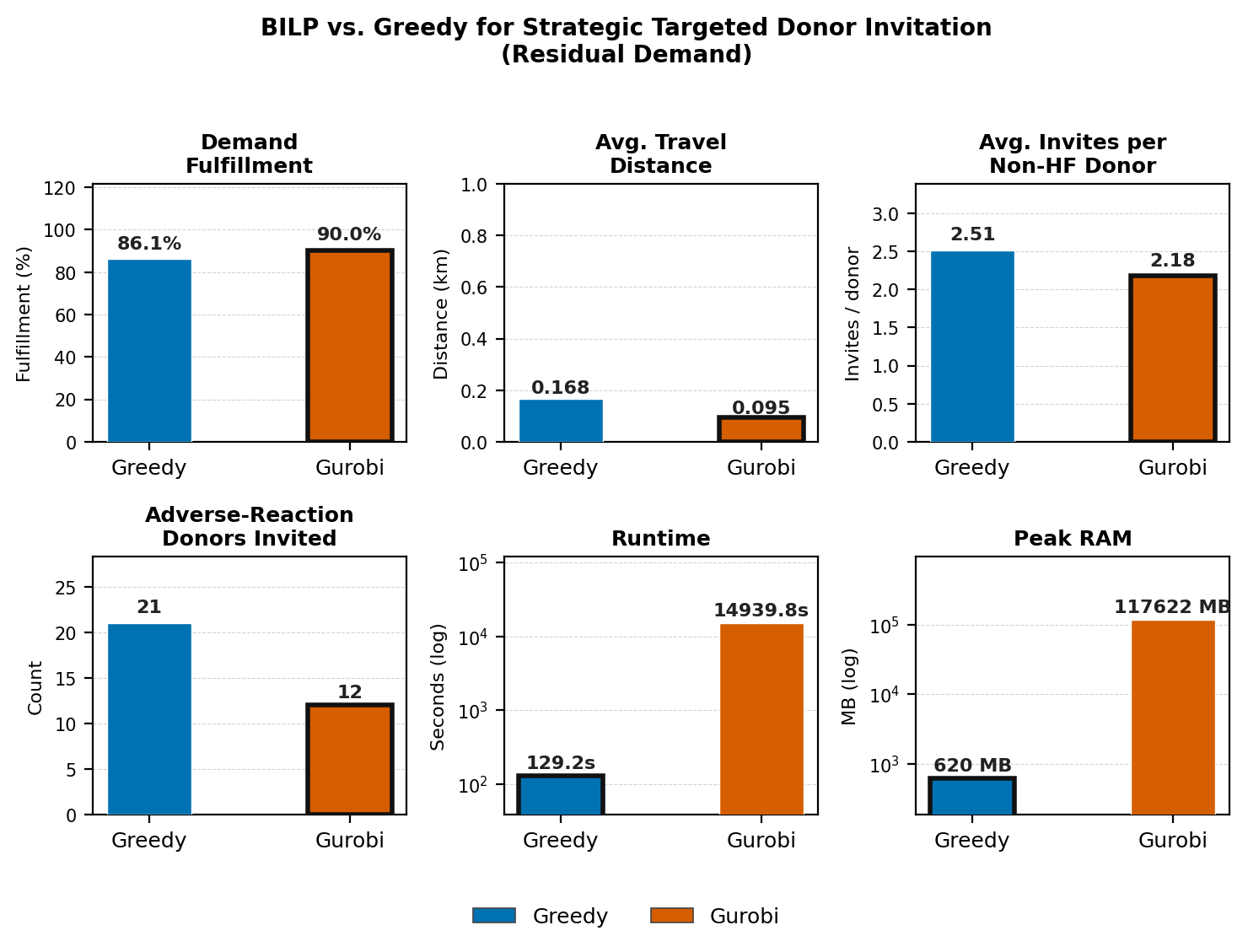}
    \caption{Comparison of solver approaches on the full Lisbon 2020 donor pool under a $2\times$ scaled residual demand scenario. \normalfont Panels show demand fulfillment rate, number of adverse-reaction donors invited, average donor--session distance (km), average invitations per non-high-frequency donor, wall-clock runtime (s, log scale), and peak process memory (MB, log scale). The highlighted bar in each panel indicates the best-performing solver. The greedy heuristic provides highly competitive fulfillment while running substantially faster and with far lower memory usage than the exact BILP benchmark.}
    \label{fig:solveralgor_comparison}
\end{figure}

Figure~\ref{fig:solveralgor_comparison} summarises the comparison across six metrics. The greedy heuristic achieved a demand fulfillment of 86.1\% versus 90.0\% for the exact Gurobi BILP—a gap of 3.9 percentage points—and invited slightly more adverse-reaction donors (21 vs.\ 12). However, this competitive solution quality is attained at a fraction of the computational cost: the greedy solver runs in 129.2\,s versus 14{,}939.8\,s for Gurobi ($\sim$115$\times$ faster) and consumes only 620\,MB of peak memory versus 117{,}622\,MB ($\sim$190$\times$ less). Average donor--session distance is extremely low across the board but slightly higher for the greedy algorithm: 0.17\,km versus 0.10\,km for Gurobi (70\% larger), because the exact solver globally minimises the assignment while the greedy makes local nearest-neighbour decisions within each demand group. Average invitations per non-high-frequency donor were 2.51 for the greedy approach compared to 2.18 for Gurobi. 

These results validate \emph{Greedy} as a robust alternative for the prospective pipeline: its strong solution quality, combined with its drastically lower computational footprint and zero commercial-solver dependency, makes it well-suited to operational deployment in settings where system resources are constrained or in settings where full-scale instances cannot be handled efficiently by Gurobi.

\section{Conclusion}
\label{sec:conclusion}

By formulating the donor--session assignment problem as a binary integer linear program and benchmarking it against an efficient greedy heuristic, we demonstrated that constraint-aware invitation planning is operationally viable even for donor pools comprising hundreds of thousands of donors and hundreds of collection sessions, while still maintaining very high geographic convenience under complex temporal eligibility rules.

Our case-oriented application of this framework to the Lisbon operational region suggests that strategic targeted personal invitations could improve the self-reliance of operational regions by narrowing the gap between realized attendance and demand fulfillment, in part by helping reactivate lapsed and inactive donors.

\subsubsection*{Limitations}
\label{sec:limitations}

While the attained results are promising, several limitations must be acknowledged. First, current geographic modeling relies on synthetic inference of donor and session locations based on postal codes and metadata. While sufficient for proof-of-concept, the lack of exact coordinates introduces noise into the distance calculations, potentially marking some invitations as ``convenient'' when they are practically infeasible, or conversely filtering out viable donors. Second, the framework depends on modeling assumptions that were not directly estimated from invitation-response data within IPST. In particular, the expected show-up behavior of personally invited donors plays an important role in both projected demand fulfillment and effective capacity usage, yet these attendance probabilities were approximated rather than calibrated from a dedicated outreach study. Third, some operational preferences are represented through heuristic design choices or soft penalties, such as the weighting assigned to prior adverse reactions and the prioritization logic used by the greedy solver. Although these choices are reasonable and operationally interpretable, different parameterizations could lead to different trade-offs between fulfillment, donor burden, travel distance, and safety-related preferences.

\subsubsection*{Future Work}
\label{sec:future_work}

To enhance the operational viability of this framework, future research will focus on three key areas.
First, a further priority for future work is the estimation of show-up probabilities following personal invitation. In the current framework, both demand fulfillment and the enforcement of effective capacity constraints depend strongly on the assumed attendance probability of invited donors. A dedicated empirical study within IPST would therefore be valuable to estimate these probabilities across donor groups and invitation settings, and to calibrate the planning model more reliably.
Second, to improve accessibility for public health services, 
open-source solvers such as SCIP or HiGHS \cite{achterberg2009scip,highs_website} 
will be benchmarked against Gurobi to determine if they can perform efficiently without commercial licensing costs. 
Third, the approximate location data used in this study may not fully represent the precise travel constraints of the donor pool. Future work should prioritize in gaining access to ground-truth geolocation data of both donor residences (or donor-specified preferred locations) and collection sites for superior 
validation of the attained geographic convenience constraints. 
Plans noncompliant with the enforced constraints can turn geographic bottlenecks more readily interpretable, enabling operations to adjust session placement to restore feasibility.

\subsection*{Funding}
This work was financed by national funds by FCT - Fundação para a Ciência e a Tecnologia, I.P., in the scope of the project LAIfeBlood+ (2024.07475.IACDC) and INESC-ID plurianual (UID/50021/2025), as well as from the Portuguese Recovery
and Resilience Plan through the Center For Responsible AI (C645008882-00000055) and the UID/PRR/50021/2025 unit.

\subsection*{Acknowledgments}
The authors thank Instituto Português do Sangue e da Transplantação (IPST) for the data provision and valuable support. We would also like to express our gratitude to Professor Daniel Pedro de Jesus Faria and Professor Arlindo Oliveira for their valuable insights, feedback, and fruitful discussions that helped shape this work.

\subsection*{Ethics declarations}
Ethical approval was obtained by Instituto Português do Sangue e da Transplantação (IPST) committee for the analysis of the anonymized dataset. The authors declare that they have no competing interests.

\bibliography{paper}
\bibliographystyle{icml2025}

\clearpage 
\appendix
\renewcommand{\thefigure}{S\arabic{figure}}
\setcounter{figure}{0}
\renewcommand{\thetable}{S\arabic{table}}
\setcounter{table}{0}

\section{Supplementary Figures and Tables}

\begin{figure}[htb]
    \centering
    \includegraphics[width=\linewidth]{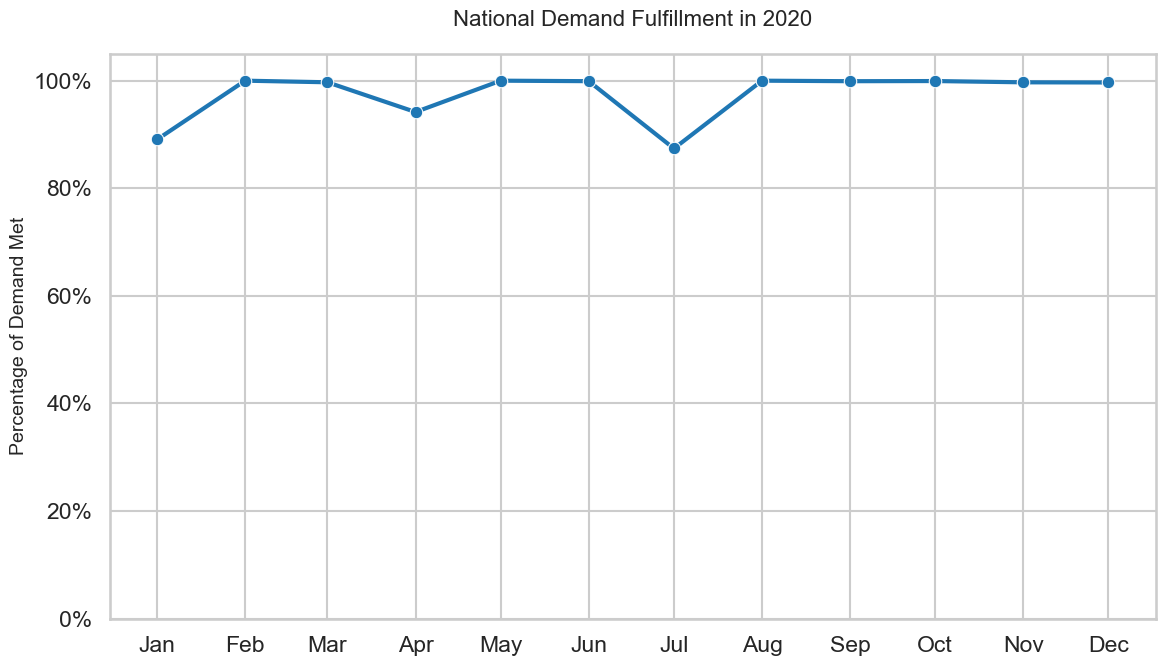}
    \caption{National demand fulfillment in 2020.}
    \label{fig:supp1}
\end{figure}

\begin{figure}[htb]
    \centering
    \includegraphics[width=\linewidth]{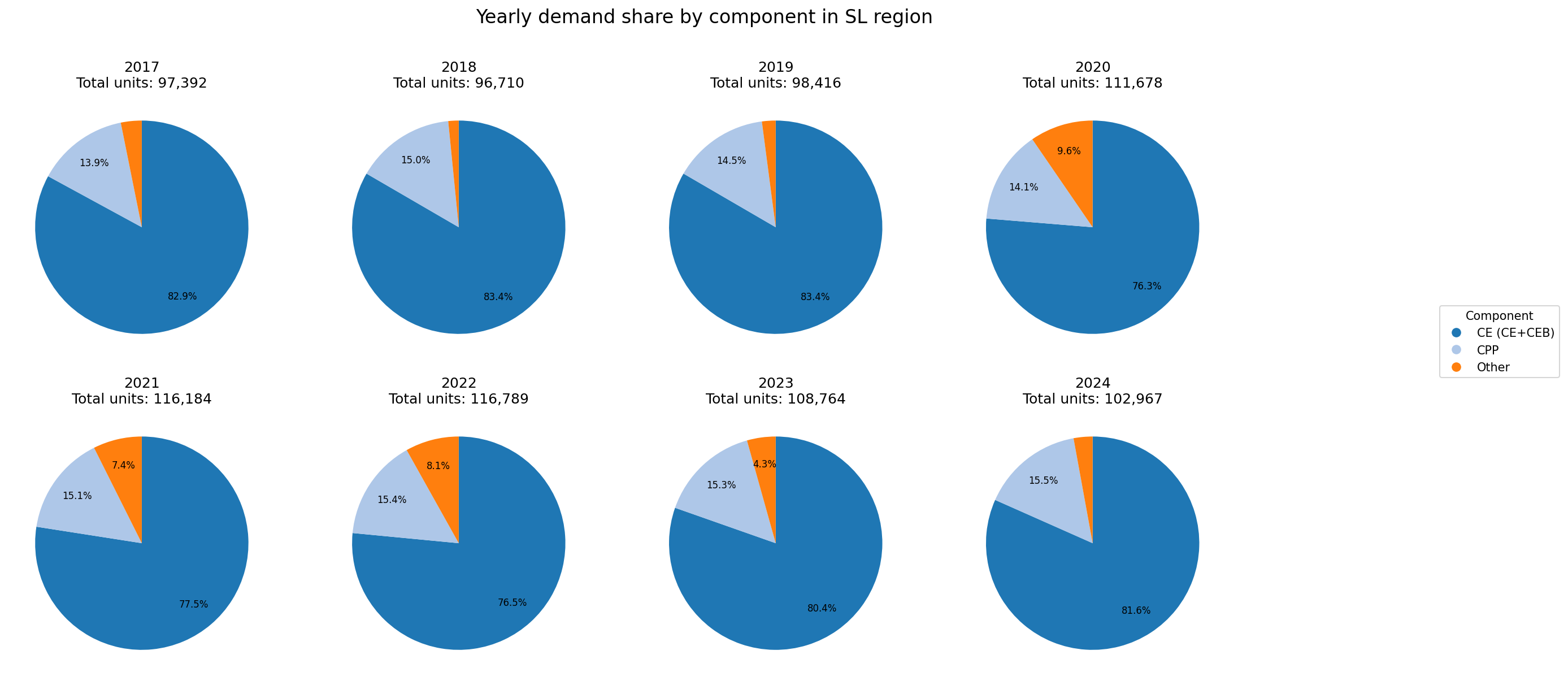} 
    \caption{Main consumed blood components in the SL operational region data.}
    \label{fig:supp_components}
\end{figure}

\begin{figure*}[t]
    \centering
    \includegraphics[width=0.62\textwidth]{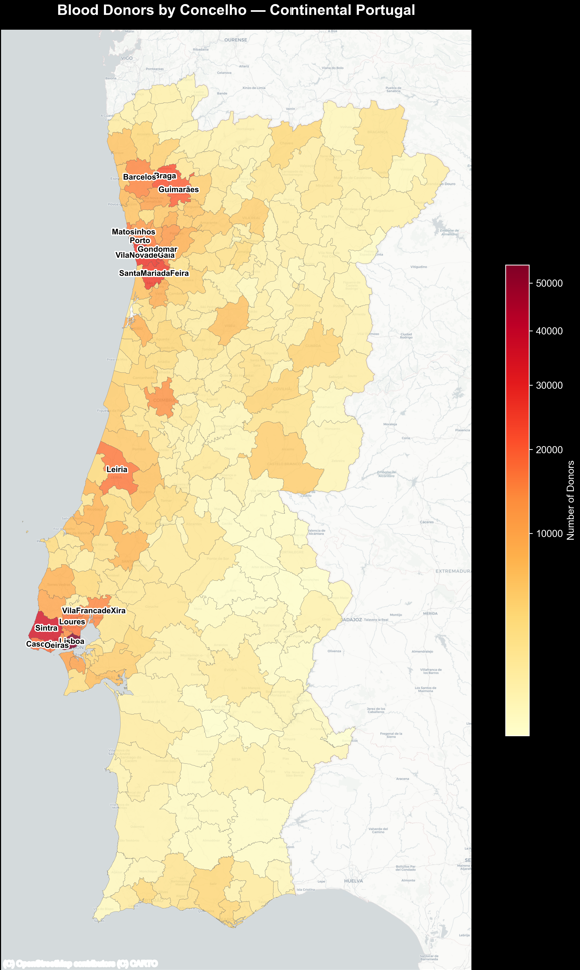}
    \caption{Distribution of Blood Donors by Municipality in Mainland Portugal. Choropleth map showing the number of blood donors across municipalities in mainland Portugal.}
    \label{fig:supp_donor_map}
\end{figure*}

\begin{figure}[htb]
    \centering
    \includegraphics[width=\linewidth]{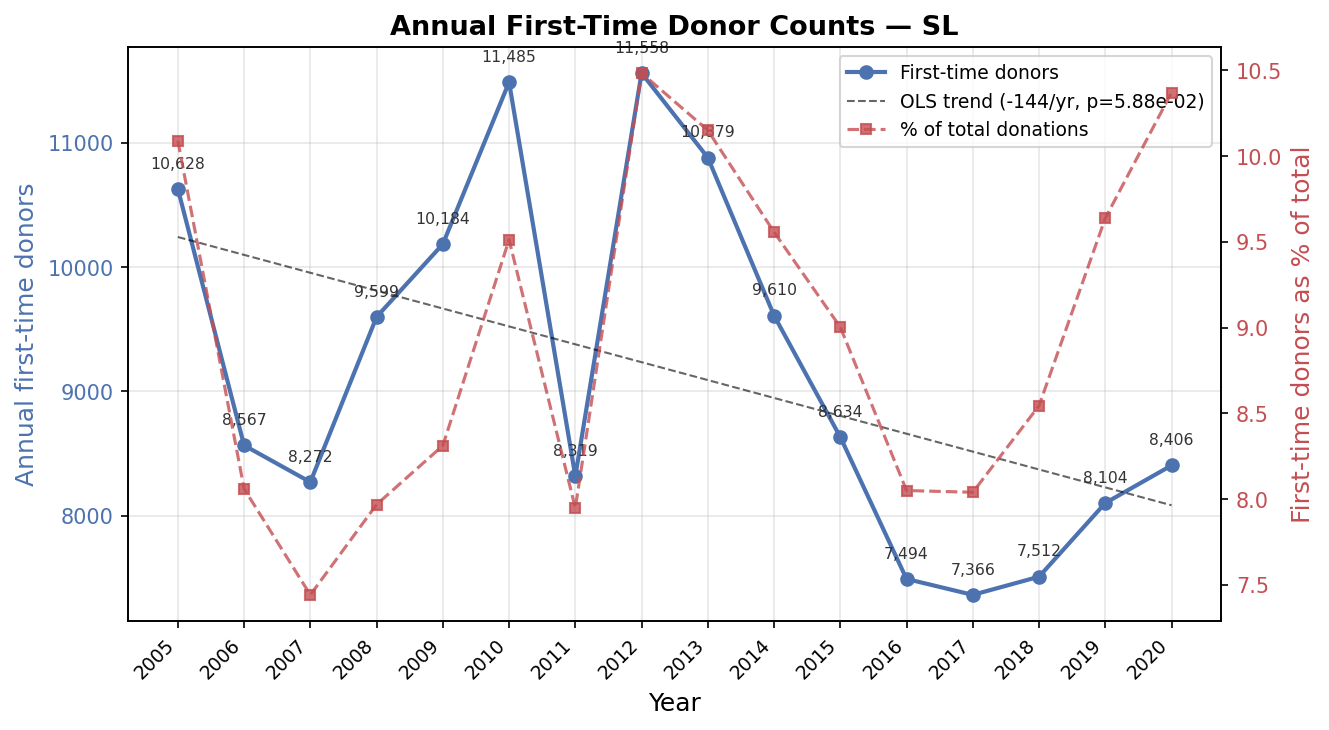}
    \caption{Annual first-time donor counts (blue, left axis) and first-time donors as a percentage of total donations (red, right axis) for the SL operational region, 2005--2020. Absolute counts declined from approximately 10{,}600 in 2005 and a peak of 11{,}500 in 2010 to 7{,}400--8{,}400 in 2016--2020, with an OLS trend of $-144$ donors per year ($p=0.059$). Despite the absolute decline, the first-time share of total donations remained relatively stable at 8--10\%, indicating that overall collection volumes have contracted in parallel. The uptick in 2020 (8{,}406 first-time donors, 10.4\% share) is consistent with pandemic-driven recruitment campaigns targeting new donor populations.}
    \label{fig:supp_annual_first_time}
\end{figure}

\clearpage 

\begin{table*}[t]
\centering
\caption{Pre-2020 training-window backtest summary for the organic attendance forecasting workflow. Mean metrics are averaged over evaluation years 2017--2020. The policy-sweep stage compared the historical baseline, logistic regression, XGBoost, and LightGBM; the MLP was introduced only in the subsequent tuning stage after the window policy had been fixed.}
\label{tab:supp_organic_backtest}
\resizebox{\textwidth}{!}{%
\begin{tabular}{llrrrrrc}
\toprule
Policy & Model & Mean Monthly MAE & Mean Relative MAE (\%) & Mean Brier & Mean ROC-AUC & Backtest Years & Best Policy For Model \\
\midrule
expanding & Historical Baseline & 8519.2 & 149.2 & 0.0839 & 0.6346 & 4 & Yes \\
rolling\_3y & Historical Baseline & 8519.2 & 149.2 & 0.0839 & 0.6346 & 4 & Yes \\
rolling\_5y & Historical Baseline & 8519.2 & 149.2 & 0.0839 & 0.6346 & 4 & Yes \\
expanding & LightGBM & 37253.8 & 652.1 & 0.1532 & 0.8373 & 4 & Yes \\
rolling\_3y & LightGBM & 38451.5 & 672.6 & 0.1609 & 0.8382 & 4 & No \\
rolling\_5y & LightGBM & 38567.0 & 675.0 & 0.1609 & 0.8382 & 4 & No \\
rolling\_5y & Logistic Regression & 423.6 & 7.6 & 0.0385 & 0.8028 & 4 & Yes \\
rolling\_3y & Logistic Regression & 433.4 & 7.7 & 0.0385 & 0.8028 & 4 & No \\
expanding & Logistic Regression & 467.8 & 8.3 & 0.0388 & 0.8012 & 4 & No \\
expanding & XGBoost & 37151.2 & 650.3 & 0.1525 & 0.8377 & 4 & Yes \\
rolling\_3y & XGBoost & 38262.3 & 669.3 & 0.1599 & 0.8385 & 4 & No \\
rolling\_5y & XGBoost & 38407.2 & 672.2 & 0.1600 & 0.8386 & 4 & No \\
\bottomrule
\end{tabular}%
}
\end{table*}

\begin{table*}[t]
\centering
\caption{Comparison of session-assignment methods for the SL operational region, 2020. Three allocation strategies are evaluated on 387 active sites: (i)~proportional historical shares, (ii)~individual preferred-site assignment (MLP + ghost-site remapping), and (iii)~the site-level GBR model. The GBR model achieves the lowest site-level MAE and median relative error, making it the preferred method for capacity estimation. The individual assignment, while less accurate at the site level, provides per-donor site linkages required by the invitation optimizer.}
\label{tab:session_methods}
\begin{tabular}{lrrrr}
\toprule
Method & \makecell{Site-level\\$r$} & \makecell{Site-level\\MAE} & \makecell{Monthly\\MAE} & \makecell{Median rel.\\err (\%)} \\
\midrule
Proportional shares                  & 0.971 & 80 & 1{,}033 & 82 \\
Individual pref.\ site               & 0.978 & 89 & 1{,}197 & 72 \\
GBR site model (reconciled)          & 0.983 & 57 & 1{,}033 & 31 \\
\bottomrule
\end{tabular}
\end{table*}

\clearpage
\end{document}